\documentclass{IOS-Book-Article}
\usepackage{ntheorem}
\usepackage{mathptmx}
\usepackage{bm}
\usepackage{mathtools}
\usepackage{soul}\setuldepth{article}
\usepackage{tabularx}
\usepackage{booktabs}
\usepackage{multirow}
\usepackage{array}
\usepackage{tikz}
\usepackage{xcolor}
\usetikzlibrary{shapes.geometric, arrows.meta, positioning, fit, backgrounds, shapes.geometric} 
\usepackage{afterpage}
\usepackage{subfig}
\usepackage{subcaption,cleveref}
\usepackage{amsmath}
\usepackage{enumitem}
\usepackage{tcolorbox} 
\usepackage{graphicx}
\usepackage{fontawesome5}
\usepackage{longtable}
\usepackage{enumitem} 
\usepackage[section]{placeins}
\usepackage{amsmath,amssymb} 
\usepackage{booktabs,tabularx,array,ragged2e,threeparttable}
\newcolumntype{Y}{>{\RaggedRight\arraybackslash}X} 
\newcolumntype{L}[1]{>{\RaggedRight\arraybackslash\hsize=#1\hsize}X} 

\usepackage{xparse}
 
\NewDocumentCommand{\runin}{O{\medskipamount} m}{%
  \par\addvspace{#1}
  \noindent\textbf{#2}\enspace%
}

\newtheorem{lemma}{Lemma}
\newtheorem{hyp}{Hypothesis}

\definecolor{lightblue}{RGB}{227,242,253}
\definecolor{lightpurple}{RGB}{243,229,245}
\definecolor{lightgreen}{RGB}{232,245,232}
\definecolor{lightorange}{RGB}{255,243,224}
\definecolor{lightpink}{RGB}{252,228,236}

\tikzstyle{process} = [rectangle, rounded corners, minimum width=3cm, minimum height=1cm, text centered, draw=black, fill=white, font=\small]
\tikzstyle{workflow} = [rectangle, rounded corners, minimum width=3.5cm, minimum height=1.2cm, text centered, draw=black, font=\small\bfseries]
\tikzstyle{agent} = [rectangle, rounded corners, minimum width=2.8cm, minimum height=1cm, text centered, draw=black, font=\small]
\tikzstyle{arrow} = [-{Stealth[length=2mm]}, thick]
\tikzstyle{dashed_arrow} = [-{Stealth[length=2mm]}, thick, dashed]
\tikzstyle{group} = [rectangle, rounded corners, draw=gray, thick, inner sep=10pt]
\tikzset{
  startstop/.style = {rectangle, rounded corners, minimum width=3.2cm, minimum height=1cm,
                      text centered, draw=black, fill=red!20},
  process/.style   = {rectangle, minimum width=4.2cm, minimum height=1cm,
                      text centered, draw=black, fill=blue!20},
  decision/.style  = {diamond, aspect=2, minimum width=3.5cm, minimum height=1cm,
                      text centered, draw=black, fill=green!20},
  arrow/.style     = {thick, -{Stealth[length=2.2mm,width=2.2mm]}}, 
}

\newcommand{\Lsetd}{\mathcal{L}_d}
\newcommand{\K}{K_d}
\newcommand{\cdf}[1]{F_{#1}}

\newcommand{\Wone}{W_1}
\newcommand{\NWone}{\widetilde W_1}

\newcommand{\DCI}{\mathrm{DCI}} 

\def\hb{\hbox to 11.5 cm{}}

\begin{document}

\pagestyle{headings}
\def\thepage{}
\begin{frontmatter}              

\title{PersonaMatrix: A Recipe for Persona-Aware Evaluation of Legal Summarization}

\markboth{}{September 2025\hb}

\author{%
  \begin{tabular}{c}
    Tsz Fung PANG\textsuperscript{a,1}, Maryam BERIJANIAN\textsuperscript{b}, Thomas ORTH\textsuperscript{a}, \\
    Breanna SHI\textsuperscript{a}, Charlotte S. ALEXANDER\textsuperscript{a}
  \end{tabular}
}
\footnotetext[1]{Corresponding Author: Tsz Fung Pang, tpang34@gatech.edu}

\runningauthor{T.F. Pang et al.}

\address[A]{Georgia Institute of Technology}

\address[B]{Michigan State University}

\begin{abstract}
 
Legal documents are often long, dense, and difficult to comprehend, not only for laypeople but also for legal experts. While automated document summarization has great potential to improve access to legal knowledge, prevailing task-based evaluators overlook divergent user and stakeholder needs. Tool development is needed to encompass the technicality of a case summary for a litigator yet be accessible for a self-help public researching for their lawsuit. We introduce \emph{PersonaMatrix}, a persona-by-criterion evaluation framework that scores summaries through the lens of six personas, including legal and non-legal users. We also introduce a \emph{controlled dimension-shifted} pilot dataset of U.S. civil rights case summaries that varies along depth, accessibility, and procedural detail as well as \emph{Diversity-Coverage Index (DCI)} to expose divergent optima of legal summary between persona-aware and persona-agnostic judges. This work enables refinement of legal AI summarization systems for both expert and non-expert users, with the potential to increase access to legal knowledge. The code base and data are publicly available in GitHub\footnotemark[2]. 
\end{abstract}
\end{frontmatter}
\markboth{September 2025\hb}{September 2025\hb}

\footnotetext[2]{\url{https://github.com/sampangtf/persona-matrix}} 
\section{Introduction}

\begin{figure}[!tbp]
    \centering
    \subfloat[Optimal Shift Level Heatmap]{\includegraphics[width=0.48\linewidth]{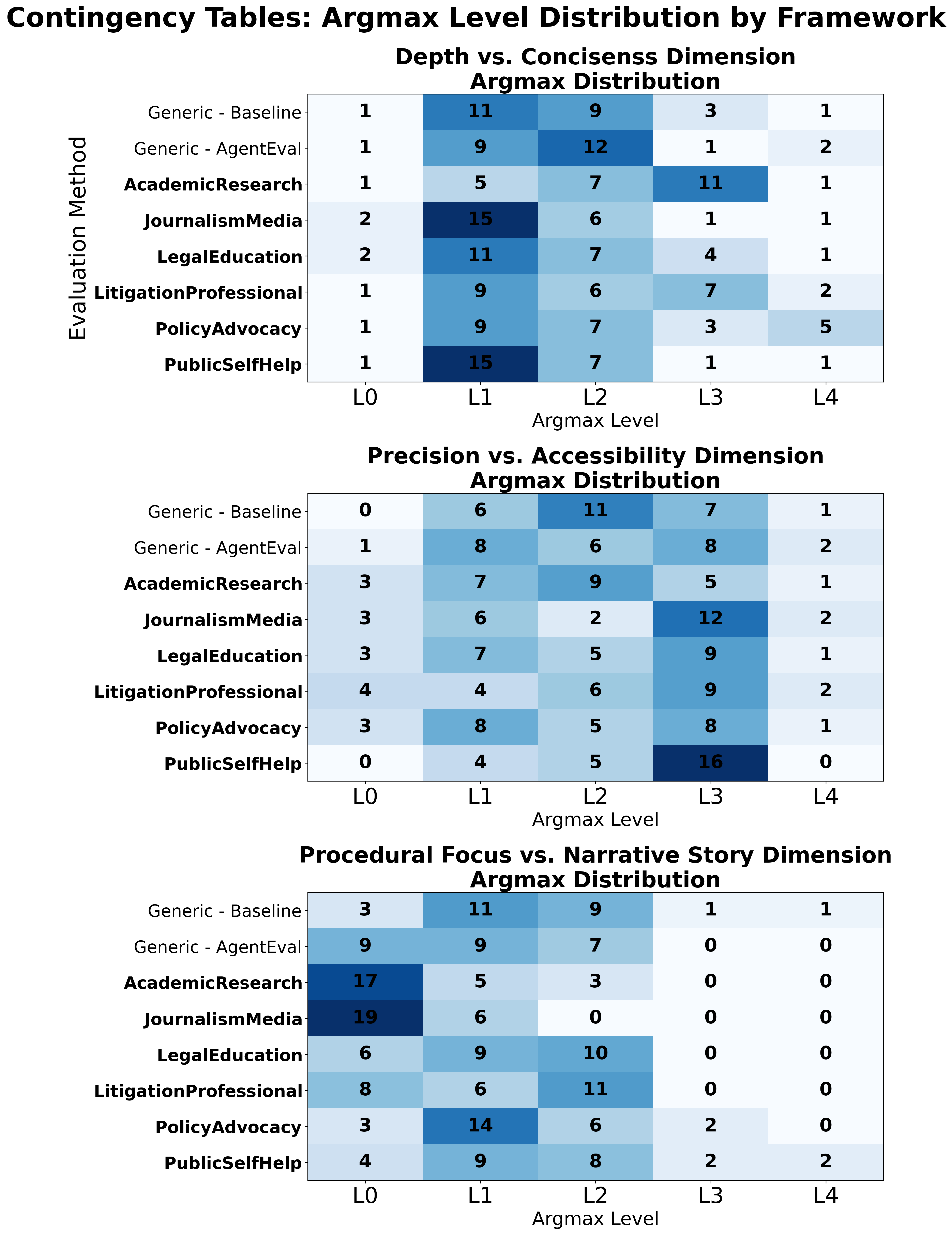}
    \label{fig:argmax-heatmap}
    }
    \hfill\subfloat[Persona Judges' Deviated Optima from Generic]
    {\includegraphics[width=0.48\linewidth]{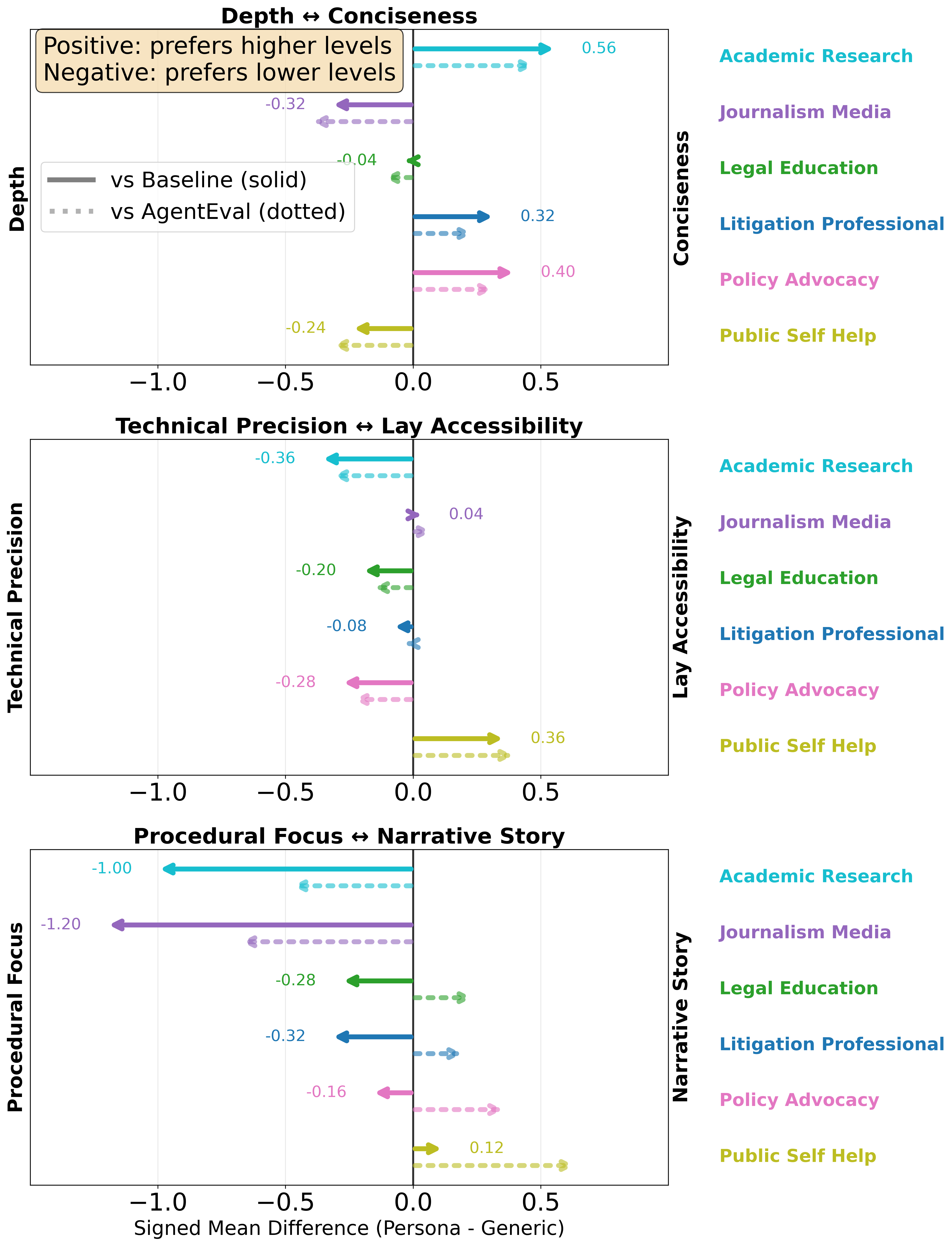}\label{fig:mean-optim-shift}
    }
    \caption{(a) Distribution of optimal (argmax) levels along three summary quality dimensions (Depth vs. Conciseness, Technical Precision vs. Lay Accessibility, and Procedural Focus vs. Narrative Story) for 6 persona judges and 2 generic judges over 25 civil rights cases. (b) Direction and magnitude of deviation of persona-aware judges' mean optimal levels from generic judges.}
    \label{fig:combined-heatmap-arrow}
\end{figure}

Automatic summarization of lengthy, complex legal documents can expand access and comprehension for diverse end users. Although large language models (LLMs) have advanced fluency and coherence, evaluating whether a summary meets user needs remains challenging. Traditional metrics such as ROUGE \cite{rouge}, BLEU \cite{bleu}, METEOR \cite{meteor}, and BERTScore \cite{BERTScore} measure lexical or semantic similarity to references, but such overlap is an unreliable proxy for utility. N-gram methods can only reward wording mimicry and even favor incoherent outputs \cite{callison-burch-etal-2006-evaluating, akter-etal-2022-revisiting}; embedding-based methods like BERTScore improve semantic matching yet still prioritize closeness to a reference over \emph{fitness for task}. 

As LLMs are deployed in domain-specific settings, LLM-as-a-judge and agentic-judge frameworks have emerged as alternatives. Structured designs such as AgentEval \cite{arabzadeh-etal-2024-assessing} stabilize judgments and improve explainability, sometimes correlating better with ground truths. Yet these methods assume a single notion of utility. In legal summarization, stakeholders have divergent priorities, and collapsing judgments into a scalar score with standardized criteria obscures real trade-offs. We therefore introduce \textbf{\textit{PersonaMatrix}}, a persona-by-criterion evaluation framework that fills the gap by capturing multi-objective utility and distinct priorities across heterogeneous stakeholders. It operationalizes persona-centered utility via scalable persona-conditioned evaluators, offering a middle ground between costly user studies and generic automated metrics.

Stakeholder-sensitive evaluation requires datasets where the same case is summarized along controlled quality dimensions. In this paper, a \textit{quality dimension} refers to a specific controllable aspect of writing that characterizes a summarization, including but not limited to lexical choice, stylistic register, compression ratio and content selection. To probe trade-offs systematically, we construct a \textbf{\textit{controlled dimension-shifted dataset}} of U.S. civil rights case summaries, where each case is summarized along conflicting quality dimensions. A scalable LLM-driven pipeline generates and validates these controlled variants from human-written seed summaries, as shown in Figure \ref{fig:combined-generator-evaluator}, enabling sensitivity analysis at manageable cost. We instantiate six personas as representative stakeholders: (1) Litigator, (2) Legal Educator, (3) Journalist, (4) Self-Help Public, (5) Academic Researcher, and (6) Policy Advocate. For example, litigators require procedural precision and citations, whereas journalists or public-facing communicators prioritize clarity, brevity, and plain language. Documentation for other persona designs, the full code base and data are available on GitHub. Gains for one group often entail losses for another: procedural detail helps lawyers but burdens general readers; simplifying language aids readability but can obscure legal reasoning, controlling rules or citations. Figure \ref{fig:combined-heatmap-arrow} shows our evaluation results over the pilot datasets, demonstrating deviated optima of persona-aware judges relative to generic judges: baseline (vanilla LLM-as-a-judge) and AgentEval.

We further propose the \textbf{\textit{Diversity-Coverage Index (DCI)}}, a bounded metric that measures how well an evaluation framework preserves between-persona diversity while maintaining within-persona coherence. DCI benchmarks persona-aware evaluation and exposes limitations of single-perspective judges. We show that persona-conditioned evaluation diverges in a statistically significant way from persona-agnostic judges, revealing systematic biases and stakeholder-specific utility gaps, and enabling multi-lens reporting and iterative refinement to make legal summaries more useful for a wide range of users with differing informational needs and levels of legal expertise, thereby supporting equal accessibility of legal knowledge.


\paragraph{Contributions.} The key contributions are:

\begin{enumerate} 
\item Formalize legal-summary evaluation as a multi-objective problem across distinct stakeholders.
\item Assemble a controlled dimension-shifted dataset of U.S. civil rights case summaries that systematically varies summaries along three conflicting quality dimensions to probe stakeholder trade-offs, namely depth $\leftrightarrow$ conciseness, technical precision $\leftrightarrow$ lay accessibility, and procedural detail $\leftrightarrow$ narrative story.
\item Present PersonaMatrix, a persona-aware evaluation framework that couples persona-conditioned judges with the dimension-shifted dataset. 
\item Introduce a novel metric, Diversity-Coverage Index (DCI), to quantify an evaluation framework's ability to capture between–persona distinctness and within-persona coherence.
\item Provide empirical comparison between persona-aware and persona-agnostic evaluators, demonstrating statistically significant divergences in stakeholder needs and revealing systematic biases in generic evaluators.
\end{enumerate}

\section{Related Work} 

Recent AI\&Law work demonstrates the importance of persona-driven, user-centered design for legal AI systems \cite{adler2023user,kolkman2024justitia}, supported by findings of wide gaps in usability difficulties between non-lawyers and lawyers \cite{Newman01092008}. Bex (2024) explicitly calls for evaluating AI legal systems with stakeholders from practice \cite{bex2024transdisciplinary}.
Hagan (2024) advocates for multi-stakeholder criteria-based evaluation for Legal Q\&A systems with the support of human user studies \cite{hagan2024measuring}. While recent legal summarization systems target plain‑language, accessible summaries for non‑lawyers~\cite{10.1145/3614407.3643700}, we still lack evaluation frameworks that assess fitness across stakeholder groups. We fill these gaps by operationalizing persona-driven evaluation for legal summarization, enabling automatic feedback for continual refinement of legal summarization systems for all stakeholders.

LLM-as-a-judge evaluates summaries with rubric-guided scoring \cite{liu2023gevalnlgevaluationusing}, but is sensitive to prompt design and exhibits well-documented positional and verbosity biases \cite{zheng2023judging}. Agentic frameworks such as AgentEval  decompose evaluation into criteria generation and scoring, improving interpretability \cite{arabzadeh-etal-2024-assessing}. However, both assume a single optimal rubric, ignoring divergent stakeholder needs in domains like legal summarization. PersonaMatrix addresses these shortcomings by accounting for distinct informational priorities across personas and expanding evaluation beyond persona-agnostic biases. 


A parallel thread in NLP and Human-Computer Interaction (HCI) has shifted system evaluation toward assessing whether outputs support decision making and task completion, aligning with usability and user experience principles \cite{blodgett-etal-2024-human, amar2005low,lai2021towards}. Although user studies remain the ``gold standard”, they are costly and slow, limiting scalability. 

In HCI, \textit{personas} serve as archetypes grounding design in concrete user needs \cite{10.1145/1463160.1463214, 10.1145/3491102.3517589}. In NLP, personas have primarily shaped dialogue generation \cite{zhang-etal-2018-personalizing}, with little attention to their role as \emph{evaluation lenses}. PersonaMatrix is intended to uncover persona-aware utility.

Analyzing stakeholder-sensitive evaluation requires datasets where the same case is summarized along controlled quality dimensions. Existing legal datasets such as MultiLexSum \cite{Shen2022MultiLexSum} and BillSum \cite{kornilova-eidelman-2019-billsum} are valuable but lack systematic dimension-shifting for summaries of the same case. MultiLexSum varies summary length, but does not capture trade-offs beyond granularity. Outside law, PersonalSum \cite{zhang2024personalsum} offers personalized news summaries conditioned on reader profiles, but this targets individual preferences rather than conflicting, role-based notions of utility. To fill this gap, we introduce a \emph{controlled dimension-shifted} dataset of U.S. civil rights cases, explicitly varying summaries along conflicting quality dimensions to expose cross-persona trade-offs.

\section{Methodology}
 
\subsection{Controlled Dimension-Shifted Dataset}


We design a scalable LLM-driven generator that transforms a seed summary into variants shifted along stakeholder-prioritized dimensions. This produces sets of summaries of the same case with systematically varied quality dimensions, avoiding bias from differences in case complexity. These controlled variants expose persona-specific preferences along dimension-shift spectra, revealing distinct and diverse optima. We test two hypotheses:

\begin{hyp}[H\ref{hyp:optimal-persona}] \label{hyp:optimal-persona}
PersonaMatrix produces distinct, illustrative evaluation results across personas, revealing multiple optima that differ in directions and magnitudes. In contrast, persona-agnostic frameworks (LLM-as-a-judge, AgentEval) collapse to a scalar score from a single perspective, as shown by the Diversity-Coverage Index (DCI).
\end{hyp}

\begin{hyp}[H\ref{hyp:nonrandomess}] \label{hyp:nonrandomess}
The diversity detected by persona-aware evaluation reflects consistent persona signals rather than random noise under corruption tests.
\end{hyp}

We identified a list of dimensions of legal summary qualities that are in conflict with each other and are prioritized by distinct stakeholders: (1) Depth vs. Conciseness, (2) Technical Precision vs. Lay Accessibility, and (3) Procedural Focus vs. Narrative Story. We chose these dimensions because they can be defined objectively and validated with measurable metrics, enabling automatic and scalable transformation and validation, and they are likely differentially prioritized by different stakeholders, manifesting the dilemma in optimizing the summary for multi-stakeholders, as shown in the rightmost column in Table \ref{table:dimensions}. The typical stakeholder alignment is intended as \emph{refutable intuition}, not fixed mapping. 


\begin{table}[t]
\centering
\begingroup
\tiny
\setlength{\tabcolsep}{3pt}
\renewcommand{\arraystretch}{1.0}

\caption{Three quality dimensions and levels. Levels are ordinal and monotone along each axis, higher levels move further toward the right-hand label. Level 0 is the expert-authored original summary; Levels 1-4 are validated rewrites from our generator. Arrows denote the direction of shift. The last row shows how each dimension shift is validated across levels.}
\label{table:dimensions}
\begin{threeparttable}
\begin{tabularx}{\linewidth}{@{}>{\centering\arraybackslash}p{8mm} Y Y Y >{\RaggedRight\arraybackslash}p{17mm}@{}}

\toprule
\textbf{Level} &
\textbf{Depth $\Rightarrow$ Conciseness} &
\textbf{Technical Precision $\Rightarrow$ Lay Accessibility} &
\textbf{Procedural Focus $\Rightarrow$ Narrative Story} &
\textbf{Typical Stakeholder Alignment\tnote{\dag}} \\
\midrule
0 & $>$500\,words; original expert summary & Legal jargon & Full procedure & Legal professionals  \\
1 & Condense sentences; prune redundancies & Professional summary & Explain rationale & Academics \\
2 & Summarize procedural cluster & Plain language with explanation & Only central legal issues & (Varies) \\
3 & Executive brief (skim-first) & Simple analogies & Outcome-first; minimal procedure & Non-legal users\\
4 & Headlines \& bullets; very short & Basic vocabulary; no jargon & Narrative human story & Public \\
\midrule
\textbf{Val metric} &
Word count falls in a stricter band than the previous level\tnote{a} &
Readability grade strictly decreases vs. previous level\tnote{b} &
Count of procedural markers decreases vs. previous level\tnote{c} &
N/A \\
\bottomrule
\end{tabularx}
\begin{tablenotes}[flushleft]
\tiny
\item[\dag] Alignment is illustrative, not prescriptive.
\item[a] We enforce monotone concision by pre-specified length bands (Level~$k$ must be strictly shorter than Level~$k{-}1$).
\item[b] Readability is computed with standard formulas. A lower grade indicates simpler prose; Level~$k$ must be strictly lower than Level~$k{-}1$.
\item[c] Procedural markers include docket/procedure cues. We count unique occurrences from a fixed list; Level~$k$ must score lower than Level~$k{-}1$.
\end{tablenotes}
\end{threeparttable}
\endgroup
\end{table}


We developed a scalable dataset generator using an \texttt{Extractor → Rewriter → Validator} agentic pipeline (Figure \ref{fig:combined-generator-evaluator}-top) that shifts summaries along the three quality dimensions while enforcing factual consistency and progressive change through a multi-step validation process, including metric-based checks and LLM-powered pairwise critiques. Rewriting prompts and validation logic are detailed on GitHub.

Seed summaries are drawn from the Civil Rights Litigation Clearinghouse\footnotemark[3], authored and validated by legal experts--law professors and law students at the University of Michigan Law School. These texts emphasize depth, technical precision, and procedural details, enabling unidirectional quality shifts along our selected dimensions. We provide an initial probe dataset of 25 cases ($\times$ 3 dimensions $\times$ 4 levels = 300  shifted summaries with 25 originals) together with code for reproducibility.

\footnotetext[3]{\url{https://clearinghouse.net/}}

\subsection{PersonaMatrix Evaluator}\label{sec:personamatrix-eval}

\begin{figure}
    \centering
    \includegraphics[width=1.0\linewidth]{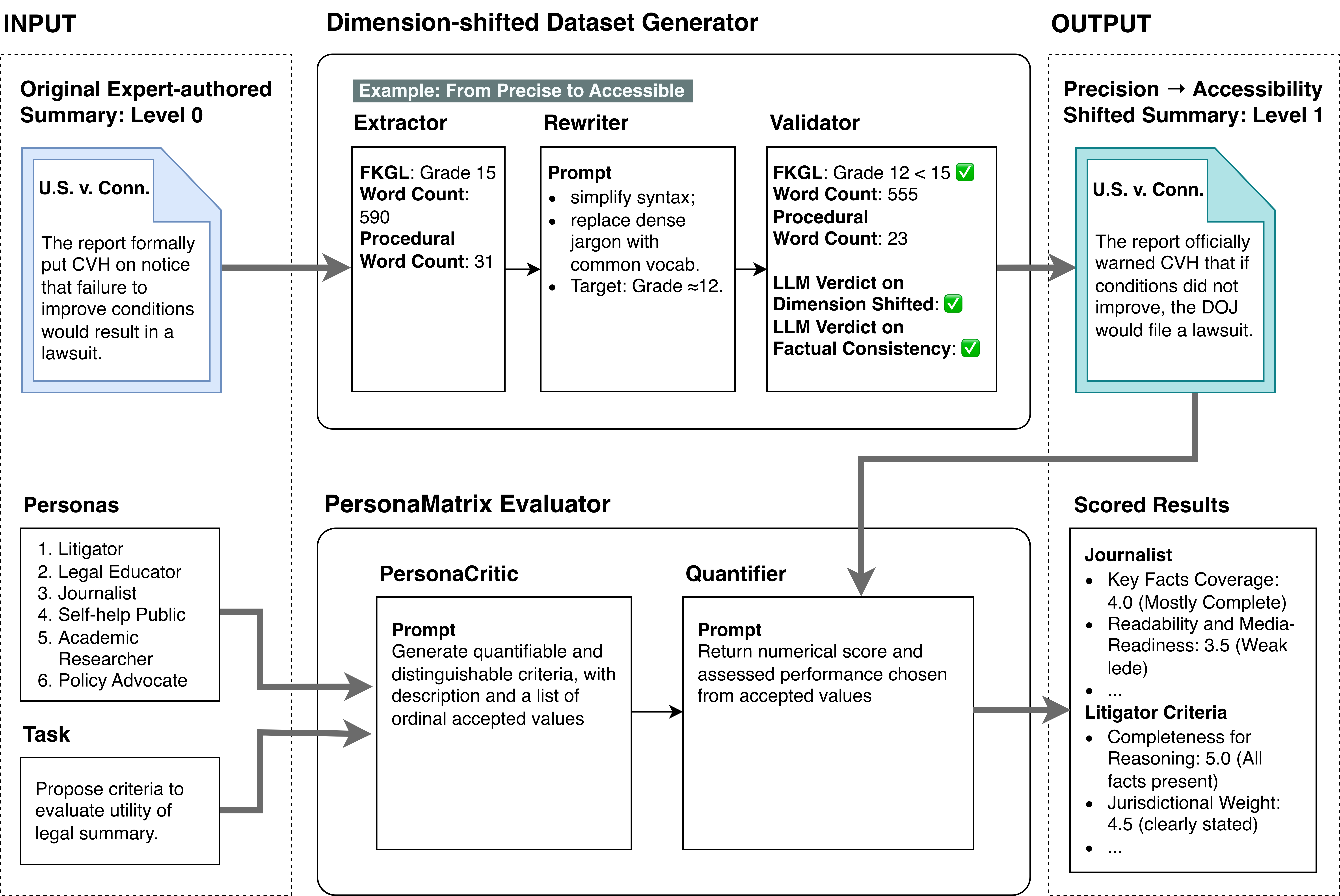}
    \caption{Demonstrations of how legal summary is rewritten in Controlled Dimension-Shifted Dataset Generator using the shift from Precision to Accessibility as example (top), and scored in PersonaMatrix Agentic Evaluator (bottom). There are $n$ PersonaCritics generating $n$ sets of evaluation criteria for $n$ personas in parallel.}
    \label{fig:combined-generator-evaluator}
\end{figure}

We adapt AgentEval's \cite{arabzadeh-etal-2024-assessing} Critic-Quantifier pipeline to a persona-aware setting. The \textbf{\textit{Critic agent}} defines task-based criteria and acceptable values; the \textbf{\textit{Quantifier agent}} scores inputs against them (Figure \ref{fig:combined-generator-evaluator}-bottom).

Let $ \mathcal{P} = \{\text{Litigators}, \text{ Public}, \text{ Journalists}, \text{ Researchers},\text{ Educators}, \text{ Advocates}\}$ denote persona judges. For each persona $p\in\mathcal{P}$, a Persona-Critic generates a rubric conditioned on persona-specific context for $p$. Each criterion $c$ defines accepted values (ordinal labels with descriptions). Full implementation can be found in the GitHub repo.

Rubrics are cached per persona and reused across cases. On top of the task-based evaluation criteria generation prompt suggested in AgentEval, we append the persona name and one-sentence  intended usage to the Critic agent's system prompt. We demonstrate that minimal context suffices to elicit stakeholder-specific criteria. Then, a Quantifier agent assigns both ordinal labels and numeric scores $s_{p,c} \in [0,5]$ for each criterion. Scores are aligned with ordinal classes (e.g., top class $\approx$ 4.5–5), requiring no rescaling. 



\subsection{Diversity-Coverage Index (DCI)}\label{sec:dci}

We propose the Diversity-Coverage Index (DCI) to quantify the structured divergence of evaluation preferences across personas, independent of the human validity of the evaluation. It is formally composed of two terms derived from the distribution of optima, which, in this work, is the argmax level of dimension shifts resulting from judges: 
\begin{enumerate}
    \item Persona signal: distinctiveness of persona-specific optima, measured by normalized mutual information (NMI). 
    \item Distinctness: deviation from persona-agnostic baselines, measured by Earth Mover's Distance (EMD) and Jensen–Shannon divergence (JSD).
\end{enumerate} 
\label{sec:dci}

\runin{Definition.} Let $d\in\mathcal{D}$  index a quality dimension with levels
$\Lsetd=\{0,\dots,\K-1\}$. Let $\K\coloneq|\Lsetd|$, $\Delta^{\K-1}\!=\{u\in\mathbb{R}_+^{\K}:\sum_{\ell=0}^{\K-1}u(\ell)=1\}$, and let $L_d$ be the level r.v. on $\Lsetd$.
$P$ denotes the persona random variable with weights $\pi=(\pi_a)$, where $\sum_a\pi_a=1$ ($\Pr[P=a]=\pi_a$).
For persona $a$, let $p_{a,d}\in\Delta^{\K-1}$ be the empirical distribution of optima (argmax over cases) on dimension $d$, i.e.\ $p_{a,d}(\ell)=\Pr[L_d=\ell\mid P=a]$. The persona-marginal probability is defined as
$q_d(\ell)=\sum_a \pi_a\, p_{a,d}(\ell)=\Pr[L_d=\ell]$.
If available, $b_d\in\Delta^{\K-1}$ serves as a persona-agnostic baseline on $\Lsetd$.
Unless otherwise stated, we use equal persona weights ($\pi=1/|\mathcal{P}|$) and set $\lambda=0.5$. Let CDF $\cdf{u} (\ell)$ be $\sum_{j=0}^{\ell}u(j)$.

\noindent The persona signal is defined as


\begingroup
\setlength{\abovedisplayskip}{-3pt}
\setlength{\belowdisplayskip}{4pt}
\setlength{\jot}{1pt}
\begin{align}
I_d &\coloneq  \mathrm{NMI}_d = \tfrac{2\,I(P;L_d)}{H(P)+H(L_d)}
   = \tfrac{2\,[\,H(q_d)-\sum_{a} \pi_{a}\, H(p_{a,d})\,]}{H(P)+H(q_d)} \in [0,1],
\end{align}
\vspace{12pt}
\noindent with joint $\Pr\,[P=a,\,L_d=\ell]=\pi_a\,p_{a,d}(\ell)$. The distinctness (persona-averaged) is

\vspace{-15pt}

\begin{align}
\bar J_d \coloneq \sum_a \pi_a\, \operatorname{JSD}_{\log_2}\!\bigl(p_{a,d}\,\Vert\, b_d\bigr) &\in [0,1],
\quad \bar S_d \coloneq \sum_a \pi_a\, \NWone\!\bigl(p_{a,d}, b_d\bigr) \in [0,1],\\
D_d &\coloneq \tfrac12(\bar J_d+\bar S_d) \in [0,1],\\
\NWone(p,b) = \Wone(p,b)/(K_d&-1),\quad
\Wone(p,b) = \sum_{\ell=0}^{K_d-1}\bigl|\cdf{p}(\ell)-\cdf{b}(\ell)\bigr|.
\end{align}

\vspace{-6pt}

\noindent The per-dimension and overall indices for $\lambda\in[0,1]$ are
\begingroup
\setlength{\abovedisplayskip}{8pt}
\begin{align}
\DCI_d(\lambda) &= \lambda I_d + (1-\lambda) D_d \in [0,1],\\
\DCI(\lambda) &= \tfrac{1}{|\mathcal{D}|}\sum_{d\in\mathcal{D}} \DCI_d(\lambda).
\end{align}
\endgroup

\vspace{-6pt}

\noindent If no baseline exists, set $\lambda=1$.

\runin{Interpretation.} $I_d$ increases when personas are internally coherent yet differ across levels. 
$D_d$ increases when persona-specific optima deviate from generic judges, both in distributional mass (JSD) and in location/consistency along the ordered scale (EMD). 

\begin{lemma}[\textbf{Shuffle Sanity}]\label{lem:shuffle}
Shuffling persona labels across i.i.d.\ cases removes persona effects on optimal levels, preserving the marginal $\pi$ independently of $\Lsetd$ but making $\tilde P\perp \Lsetd$. Thus $I(\tilde P;\Lsetd)=0$ and $I_d=0$, so $\mathrm{DCI}_d(\lambda)=(1-\lambda)D_d$. Moreover, the plug-in estimator satisfies $\widehat{\mathrm{NMI}}_d\xrightarrow{p}0$ as $n\to\infty$, where $n$ is the total number of persona–case pairs $(P,\Lsetd)$ in the shuffled sample. Proof can be accessed on GitHub.
\end{lemma}

\section{Experimental Setup}
We evaluate summaries in the controlled dimension-shifted dataset using three evaluators: (1) vanilla LLM-as-a-judge (baseline), (2) AgentEval \cite{arabzadeh-etal-2024-assessing}, and (3) PersonaMatrix. For criteria generation, we compared several reasoning LLMs (GPT-5\cite{openai_gpt5_system_card_2025}, GPT-5-mini\cite{openai_gpt5_system_card_2025}, Gemini-2.5-pro\cite{comanici2025gemini25pushingfrontier}, Gemini-2.5-flash\cite{comanici2025gemini25pushingfrontier}) and selected GPT-5-mini. 
Criteria lists are available in our GitHub.

For quantification, we use public leaderboard standings which motivate this choice: GPT-5-mini (minimal-reasoning) is preferred for its stability, discriminative power (ability to separate systems in evaluation), human correlation in creative judgment, and cost-efficiency, as evidenced by its top-5 placement on EQ-Bench Judgemark-v2.1\footnotemark[4]. To reduce bias and improve reproducibility, we blind the Quantifier to shift levels and run three stochastic passes per summary, averaging the scores. We do not scale numerical scores, since our analysis only relies on the ordinal order of scores, converting them into argmax shift levels for each legal case. 

\footnotetext[4]{EQ-Bench Judgemark v2.1 Leaderboard, assessed on 2025-08-09. Archived copy: Wayback Machine snapshot (\url{https://web.archive.org/web/20250809161133/https://eqbench.com/judgemark-v2.html}).} 

\begin{figure}
  \centering
  \subfloat[DCI Components–Procedural Dimension]{\includegraphics[width=0.48\linewidth]{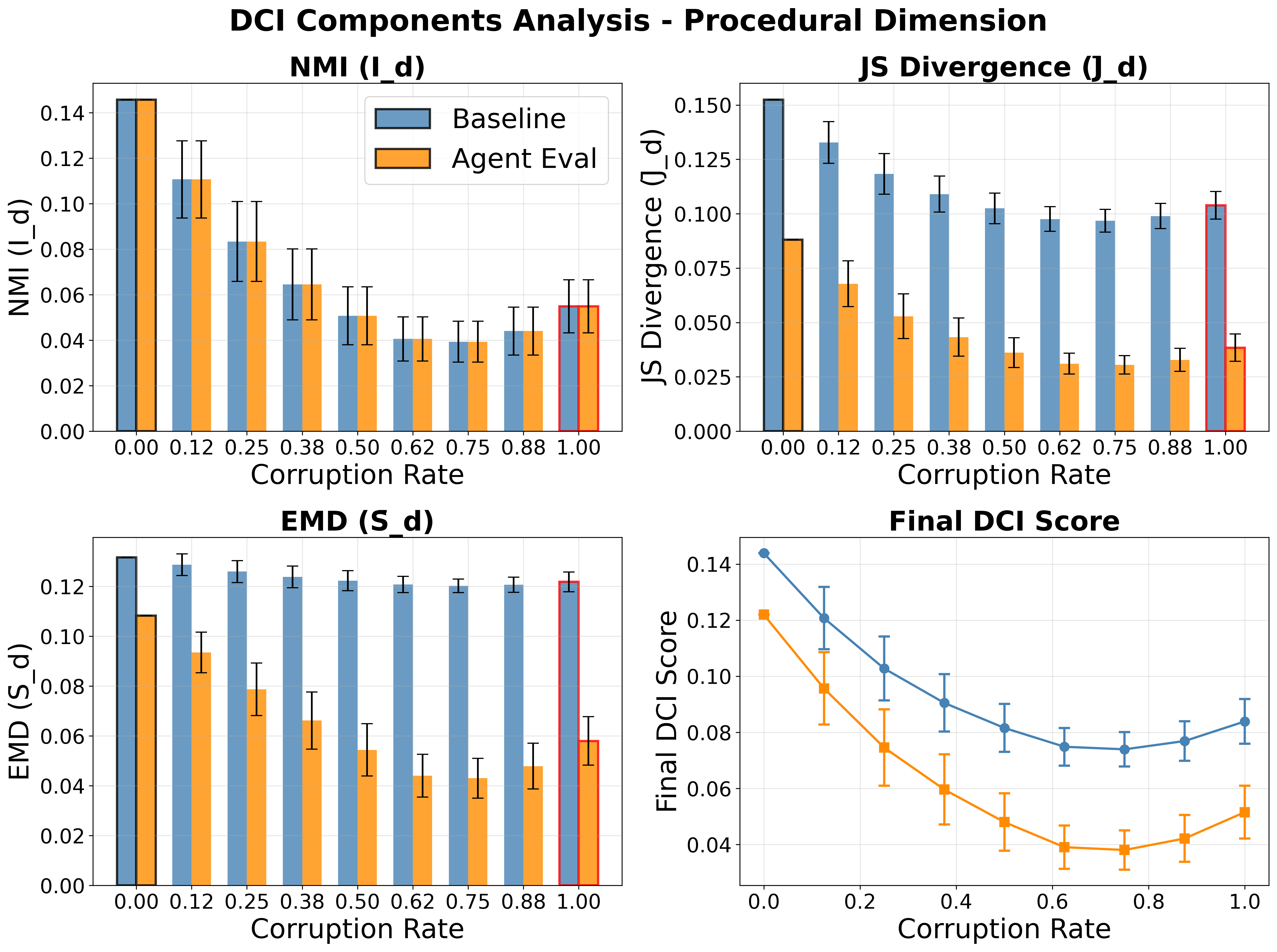}\label{fig:dci-components}}
  \hfill
  \subfloat[DCI Against Corruption Rate]{\includegraphics[width=0.48\linewidth]{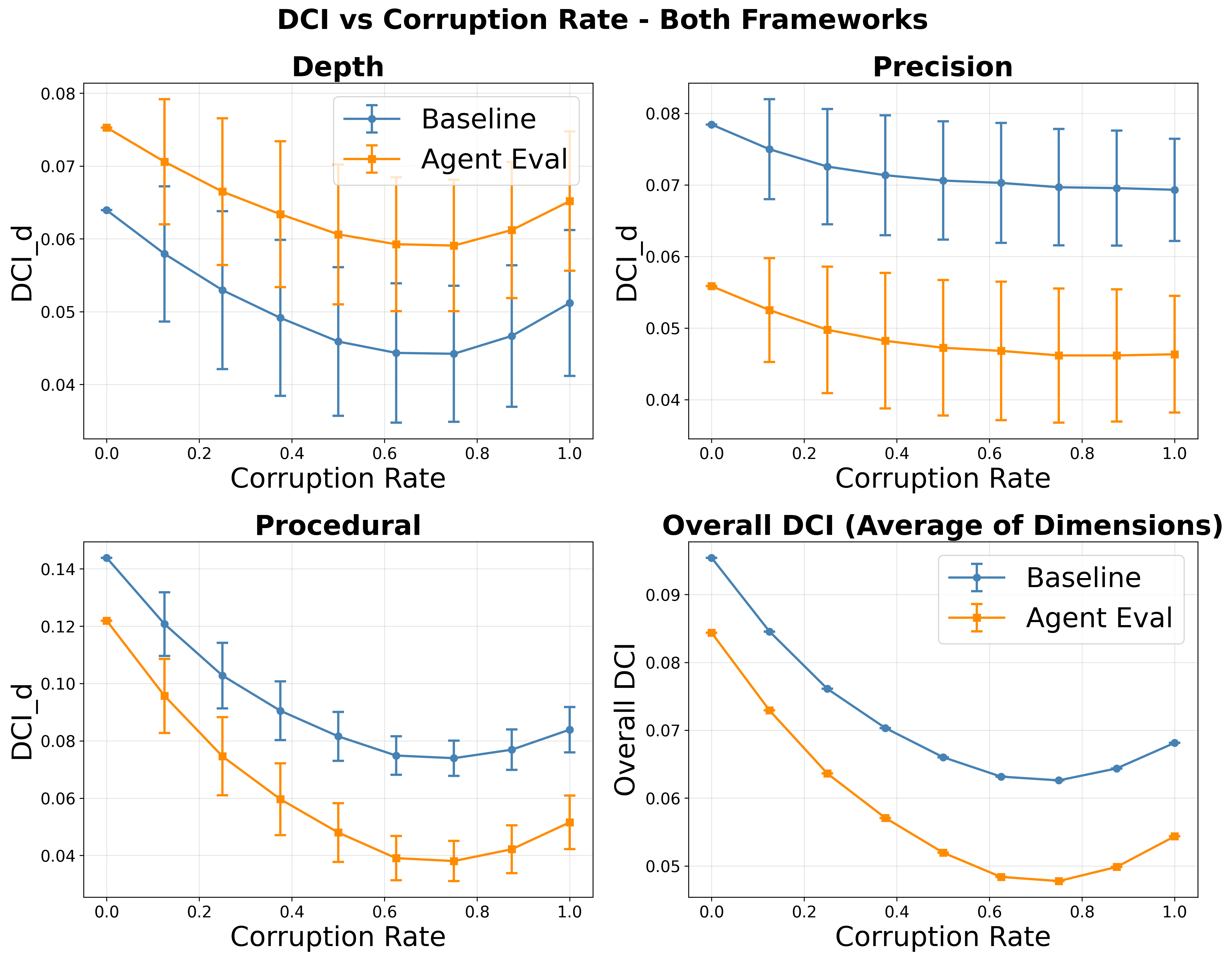}\label{fig:dci-combined}} 
  \caption{(a) The changes of DCI broken down into components against the corruption ratio over Procedural dimension, (b) changes of final DCI against increasing corruption ratio. Bottom-right figure in (a) is the same as bottom-left figure in (b).}
  \label{fig:dci}
\end{figure}


To study the impact of persona awareness, we test the robustness of PersonaMatrix against corrupted persona labels. We conduct a \textbf{\emph{corruption sweep}} by randomly permuting persona labels for a fraction $r \in [0,1]$ of cases. Each corruption ratio is repeated 2000 times for statistical stability. This allows us to track how DCI degrades as the persona signal is intentionally weakened. We report mean DCI scores with standard deviations as error bars (Figure \ref{fig:dci}). 

\section{Results and Discussion}

\paragraph{\textbf{Persona-aware judges exhibit divergent optima.}}

We convert numerical scores to argmax shift levels per case, and summarize the resulting distributions in Figure \ref{fig:argmax-heatmap}. In the case of ties, we use the median rounded level with the max scores as the argmax. Relative to generic judges, persona-conditioned judges exhibit shifted optima in both direction and magnitude across all three dimensions (summarized in Figure \ref{fig:mean-optim-shift}). Concretely:

\begin{itemize}
    \item Depth vs. Conciseness. Persona optima concentrate at intermediate levels (L1--L2), rather than the originals (L0), indicating that ``more detail'' is not universally preferred. This supports our multi-objective view of legal summarization: informative yet skim-efficient abstracts are often optimal for downstream users.
    \item Technical vs. Accessible. Journalists and public-facing personas tilt toward higher accessibility (L3), whereas legally oriented personas prefer more technical phrasing. Extreme accessibility (L4) is disfavored across personas, consistent with criteria that reward both clarity and retention of dispositive content.
    \item Procedural vs. Narrative. Most personas avoid highly narrative variants (L4), but ``self-help" shows mild tolerance for L2--L4 when factual outcomes and stakes remain salient. 
\end{itemize}


\paragraph{\textbf{Statistically significant divergence between persona-aware and generic judges.}} These shifts are not just qualitative. The $\chi^2$ tests show statistically reliable divergences from the generic judges in several persona-dimension pairs: Journalism/Media differs from Baseline on Precision $\leftrightarrow$ Accessibility (\(p{=}0.028\)) and Procedural $\leftrightarrow$ Narrative (\(p{=}0.001\)); Academic/Research differs from Baseline on Procedural $\leftrightarrow$ Narrative (\(p{=}0.002\)) and from AgentEval on Depth $\leftrightarrow$ Conciseness (\(p{=}0.025\)). These results empirically support Hypothesis H\ref{hyp:optimal-persona}: persona-aware judgments expose trade-offs missed by single-perspective judges.



\paragraph{\textbf{Operationalizing DCI: persona signal and deviation from generic judges.}}

Using the argmax distributions as inputs, we compute the components of the DCI (Section \ref{sec:dci}). Before corruption ($r{=}0$), PersonaMatrix shows high persona signal (NMI) and clear deviations from persona-agnostic judges (JSD, EMD). As we gradually corrupt persona labels (shuffle rate \(r\in[0,1]\)), the components decay almost monotonically across three dimensions. Figure \ref{fig:dci-components} shows changes in DCI components for Procedural $\leftrightarrow$ Narrative dimension. Across three dimensions, the combined DCI drops accordingly (Figure \ref{fig:dci-combined}). This matches Lemma \ref{lem:shuffle}: shuffling destroys persona-level associations, driving $\mathrm{NMI}\!\to\!0$ and reducing distinctness from generic judges. These trends support Hypothesis H\ref{hyp:nonrandomess}: diversity signals are not artifacts of noise.


\paragraph{\textbf{Why does Precision $\leftrightarrow$ Accessibility decay more slowly?}} Two factors likely compress separation on this axis (Figure \ref{fig:argmax-heatmap}). First, the accessibility rewrite deliberately preserves length and procedural cues, dampening gains from persona conditioning relative to the other axes. Second, readability constraints induce mid-level convergence: several rubrics reward plain phrasing and preservation of dispositive content, which tends to select L2--L3 for multiple personas, yielding flatter decay under corruption.

\paragraph{\textbf{High-corruption rebound is a clustering artifact.}} The small \(\mathrm{DCI}\) uptick at high-corruption level (as \(r\) increases from 0.75 to 1) shown in Figure \ref{fig:dci-combined} is consistent with random re-assembly of partial persona clusters during permutation. Personas with similar distribution of optima towards shift levels could be re-assigned to the same pseudo-persona, thus regaining partial persona-aware signals. For example, legal educators and litigators share similar optima patterns and may be clustered again upon near-full shuffle of persona labels. 

\paragraph{\textbf{Linking trends to criteria: why persona-aware beats generic?}}
The divergence patterns align with the learned rubrics. For instance, generic AgentEval includes ``Completeness of Core Case Elements", which implicitly rewards procedural density and can bias toward L0--L1 on Depth $\leftrightarrow$ Conciseness, consistent with generic peaks at lower shift levels. In contrast, litigation-focused criteria such as ``Holding clarity and citable form" and ``Clarity, Conciseness,
Formatting, and Navigability" value rule-like, extractable statements; this rationalizes Litigator preferences for mid-level concision and moderate accessibility (L1--L2/L3). On the narrative axis, Litigator's ``Material Facts—Relevance and Completeness" and AgentEval's ``Brevity vs. Necessary Detail" penalize story-heavy rewrites, explaining the broad disfavor of L4.

\paragraph{\textbf{Takeaways for evaluation practice:}}
\begin{itemize}
    \item Intermediate optima are the norm. Across personas, the best summaries are typically not the longest or the simplest; they sit at interior points on each axis, reinforcing that legal summarization is a multi-objective balancing problem.
    \item Persona lenses change the answer. Statistically significant $\chi^2$ differences and high DCI at $r{=}0$ show that one rubric cannot faithfully stand in for heterogeneous stakeholders.
    \item Corruption tests matter. The near-monotone DCI decay under label shuffling is a strong, scalable sanity check that persona signals are real rather than prompt artifacts.
\end{itemize}


\section{Limitations and Future Work}

PersonaMatrix is an LLM-driven pilot for surfacing stakeholder-conditioned preferences, but it remains an early proof of concept rather than validated stakeholder evidence. The current controlled dimension-shifted dataset is small, synthetic, and U.S.–specific, and the personas are coarse approximations of real stakeholder diversity. Consequently, the statistical evidence is illustrative and fragile, limited by sample size, lack of multiple-testing correction, and simplifying assumptions such as argmax scoring. Future studies should expand the dataset’s size and domain, integrate human-annotated baselines, and test the independence of the three quality axes, since correlations among them may confound persona effects. These steps would strengthen both the robustness and generalizability of the findings.

The DCI measures systematic divergence among personas but does not capture human validity or satisfaction. Likewise, persona optima derived from agentic judges (Figure \ref{fig:argmax-heatmap}) do not always align with intuitive stakeholder mappings in Table \ref{table:dimensions}, emphasizing the difficulty of defining persona-specific utility. Future work should validate persona criteria and optima through stakeholder surveys that directly assess whether LLM-based evaluations align with perceived usefulness for actual users.


Rubric reliability also remains untested. Rerunning rubric generation and quantifying its stability would improve reproducibility. Some LLM-generated criteria (e.g., metadata or link checks) can be spuriously satisfied or double-count holdings, encouraging verbosity. While macro-averaging and corruption tests mitigate this, future work should detect and down-weight low-observability criteria.



Finally, the approach relies entirely on LLMs to generate summaries, define criteria, and serve as evaluators, without direct stakeholder input. For a truly stakeholder-oriented framework, human participation is essential to evaluate relevance, clarity, and utility. Establishing this external validation is the critical next step.

\section{Conclusion}

Legal summarization is inherently multi-objective, demanding evaluation frameworks that capture diverse notions of utility. 
PersonaMatrix addresses this by combining controlled dimension shifts, persona-conditioned criteria, and the Diversity-Coverage Index (DCI) to reveal how stakeholder perspectives diverge. Across all axes, persona-aware judges select interior optima that differ by stakeholder, whereas persona-agnostic judges collapse toward single, generic peaks. These patterns (high NMI, non-trivial JSD/EMD, and decay under label corruption) indicate systematic, non-random persona signals.

PersonaMatrix offers a practical recipe for stakeholder-sensitive evaluation in high-stakes settings. In practice: (1) one-size-fits-all judges risk mis-scoring summaries tailored to different users; (2) reporting persona-specific and generic results together exposes real trade-offs; and (3) DCI provides a compact indicator of between-persona distinctness and within-persona coherence.

Despite current scale and validation limits, PersonaMatrix shows that evaluation frameworks and, by extension, legal AI systems benefit from multiple stakeholder lenses rather than a single universal rubric. It lays groundwork for human-validated, stakeholder-aware evaluation pipelines that make AI-generated legal summaries more accurate, accessible, and socially relevant for both experts and the public.



\bibliographystyle{vancouver}
\bibliography{references}
\end{document}